\documentclass[10pt,letterpaper]{article}

\usepackage{cogsci}

\cogscifinalcopy % Uncomment this line for the final submission 

\usepackage{pslatex}
\usepackage{apacite}
\usepackage{graphicx}
\usepackage{xcolor}
\usepackage{amsmath}
\usepackage{subcaption}
\usepackage{multicol}
\usepackage{amsfonts}

\usepackage{float} % Roger Levy added this and changed figure/table
\title{Distilling Symbolic Priors for Concept Learning into Neural Networks}
 
\author{{\large \bf Ioana Marinescu,\textsuperscript{1} R. Thomas McCoy,\textsuperscript{2} Thomas L. Griffiths\textsuperscript{1,3}} \\
  \texttt{ioanam@princeton.edu}, \texttt{tom.mccoy@yale.edu}, \texttt{tomg@princeton.edu}\\
  \textsuperscript{1}Department of Computer Science, Princeton University \\
  \textsuperscript{2}Department of Linguistics, Yale University \\
  \textsuperscript{3}Department of Psychology, Princeton University \\
}

\begin{document}

\maketitle

\begin{abstract}
Humans can learn new concepts from a small number of examples by drawing on their inductive biases. %that support abstraction and generalization. 
These inductive biases have previously been captured by using Bayesian models defined over symbolic hypothesis spaces.
%Bayesian models defined over symbolic hypothesis spaces have previously been used to capture these inductive biases. 
Is it possible to create a neural network that displays the same inductive biases? 
%Doing so would extend the insights from these Bayesian models to a new---and more cognitively plausible---implementation, one that is based on parallel distributed processing rather than explicit symbolic representations. 
%But does this mean that capturing human learning requires explicit symbolic representations? 
%However, standard neural networks require far more training data than humans do, limiting the extent to which they can serve as models of human learning. 
%One type of model that does not use explicit symbolic structure is neural networks, but standard neural networks require far more training data than humans do, limiting the extent to which these systems can serve as models of human concept learning. 
%Neural networks are powerful concept learners but they require a lot of training data, unlike humans. 
We show that inductive biases that enable rapid concept learning can be instantiated in artificial neural networks by distilling a prior distribution from a symbolic Bayesian model via meta-learning, an approach for extracting the common structure from a set of tasks. %, resulting in a system that has inductive biases aligned with those tasks. By generating the set of tasks used in meta-learning from the prior distribution of a Bayesian model, we are able to transfer that prior into a neural network. 
We use this approach to create a neural network with an inductive bias towards concepts expressed as short logical formulas. Analyzing results from previous behavioral experiments in which people learned logical concepts from a few examples, we find that our meta-trained models are highly aligned with human performance. %For instance, in one experiment a meta-trained model had a 0.98 correlation coefficient with humans, while a randomly initialized model had 0 (3.467e-16). 
%These results show that meta-learning can be used to bridge the divide between probabilistic modeling and neural networks to instantiate structured inductive biases in connectionist models. %combine the strengths of probabilistic modeling, symbolic representations, and neural networks.

\end{abstract}

\section{Introduction}
\begin{figure}[b]
    \centering
    \includegraphics[width=\linewidth]{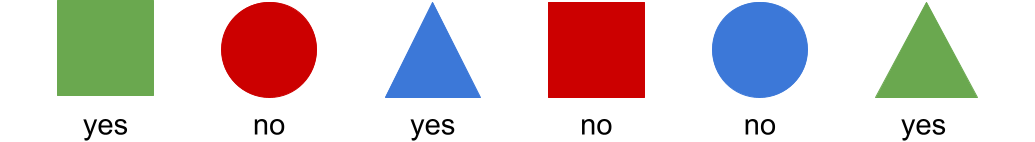}
    \caption{Concept learning from examples. The concept underlying the labels is \textit{green or triangle}: objects are labeled \textit{yes} if they are \textit{green} or a \textit{triangle} and \textit{no} otherwise. In our experiments, learners are given a set of labeled examples such as these and are then required to predict the labels for an additional set of examples.}
    \label{fig:green-or-triangle}
\end{figure}

People can make rich inferences from remarkably little data. 
%perform various tasks effectively and efficiently even without much prior experience. 
For instance, consider the domain that we focus on in this work: concept learning. People can learn a new concept from just a few examples (e.g., \citeNP{bloom2002children,xu2007word,lake2020people}). Figure \ref{fig:green-or-triangle} illustrates the concept \textit{green or triangle}; the objects are labeled \textit{yes} if they are an instance of this concept and \textit{no} otherwise. 
Given such data, people can rapidly infer what concept underlies the labeling, for a wide range of concepts \cite{bruner1956study,feldman2000minimization,goodman,piantadosi}. 
%In this particular case, some inferences might include: \textit{not circle}, \textit{not blue}, or \textit{not red}, since all circles, all blue, and all red objects are labelled \textit{no}.

Bayesian models have successfully been used to capture the human ability to learn from few examples \cite{goodman, piantadosi}. Although such models are effective at explaining human learning behavior, they are often built on explicitly symbolic hypotheses, which have been argued against as components of mechanistic cognitive theories by advocates of artificial neural networks \cite{mcclelland2010letting}. Recent neural network models 
powered by deep learning \cite{lecun2015deep} have been shown to be extremely effective at solving a variety of problems, including learning to classify stimuli into categories. For instance, for the popular ImageNet classification challenge \cite{imagenet}, the best-performing systems are neural networks (e.g., \citeNP{yu2022coca,chen2023pali}).
However, the neural networks that are used for these tasks typically require far more training examples than humans need. For instance, ImageNet contains approximately 1200 training examples per class. When standard neural networks are instead trained on smaller amounts of data, they often generalize poorly \cite{lake2015human,hoiem2021learning}.

Can we use the symbolic representations from Bayesian models to enable neural networks to acquire concepts from smaller amounts of data? Answering this question is important in order to evaluate the potential of artificial neural networks as models of human learning, providing a path towards identifying more plausible cognitive mechanisms that are informed by Bayesian modeling.
This problem is at its core about \textit{inductive biases}---the factors that guide how a learner generalizes. Bayesian models are effective at capturing human inductive biases \cite<e.g.,>{goodman,piantadosi}. Therefore, if neural networks can be given the same inductive biases as Bayesian models, then they could match human learning behavior as effectively as Bayesian models do. 
How, then, can a Bayesian model's inductive biases be given to a neural network?

%The natural tendency to think about information in a structured way allows us to make connections and generalize to new situations more easily \cite{?}. The human ways of thinking are different from the learning processes of neural network-based machines, which do not have the inherent biases that we do, enabling us to learn efficiently. Previous work has been exploring ways to identify and incorporate these biases into artificial intelligence systems to improve their ability to learn and generalize in a way that is more similar to human learning \cite{?}.

%\textcolor{red}{Briefly allude to why LLMs don’t solve the problem - to be discussed more in Discussion section}

\begin{figure}[t]
    \centering
    \includegraphics[width=\linewidth]{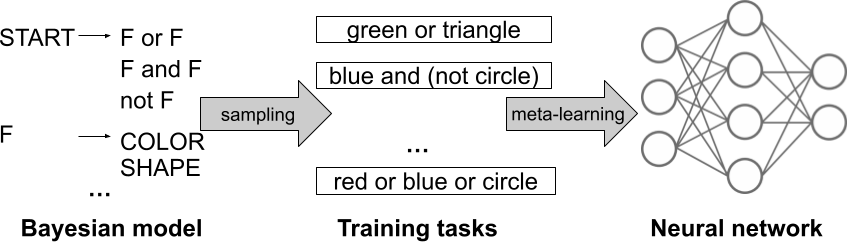}
    \caption{Learning a prior from a Bayesian model and distilling it into a neural network}
    \label{fig:method}
\end{figure}

In this work, we use meta-learning to distill the inductive biases from a Bayesian model into a neural network. 
%This approach has two key components: Bayesian modeling and meta-learning. 
The Bayesian model serves as our characterization of human inductive biases; in the Bayesian framework,
%One approach that has been used to capture human learning of abstract concepts from small amounts of data is to characterize this problem as one of Bayesian inference \cite{?}. Using this probabilistic modeling approach, human 
an inductive bias can be characterized as a prior probability distribution over hypothesized concepts. 
%This Bayesian approach has been highly successful at modeling human concept learning from small numbers of examples \cite<e.g.,>{goodman,piantadosi}, motivating its use in our framework as a way to characterize human biases.
Meta-learning is then used to turn this prior into a training regime for a neural network. 
%A promising technique for endowing neural networks with inductive biases is through the use of meta-learning  \cite{metalearning, metareasoning}. 
In meta-learning, a system is trained on many related tasks in order to learn the underlying abstractions that are common across these tasks, thereby giving it inductive biases that enable it to readily learn new tasks \cite{thrun, schmidhuber:1987:srl}. 
In our application of meta-learning, each task is a concept sampled from our Bayesian prior, such that meta-learning should result in the neural network internalizing this Bayesian prior.
%Usually the set of tasks used in meta-learning is determined by sampling from the same task-distribution. We adapted this approach to distill the prior from a Bayesian model by sampling tasks representing logical concepts from probabilistic context-free grammars. 
In this way, meta-learning serves as a formal framework for bridging probabilistic models of human concept learning and deep neural networks (see Figure \ref{fig:method}).

In our experiments, we used this approach to distill the prior from the Rational Rules model \cite{goodman} into a neural network. Across a range of previous behavioral experiments, the resulting neural network displayed generalization behavior that was highly aligned with both humans and the Rational Rules model. In contrast, a standard neural network fared poorly across these conditions. 
These results demonstrate that meta-learning can successfully be used to integrate Bayesian and connectionist models of concept learning, enabling the creation of models that combine the complementary strengths of these two approaches.
%We studied a series of behavioral experiments regarding logical concept learning from prior work and found that models initialized with the Bayesian prior from meta-learning are highly aligned with humans and greatly outperform the randomly initialized models. 

\section{Background}
 % Should cover (i) Bayesian concept learning setup (move probabilistic model stuff below up here)
 % (ii) neural networks and prior connectionist approaches to few-shot learning,  and (iii) details of setup for meta-learning and MAML in particular, plus references to its uses in cognitive science (Tom M's work, Marcel Binz review).

\subsection{Bayesian concept learning}
\begin{figure}
  \centering
  \includegraphics[width=0.5\textwidth]{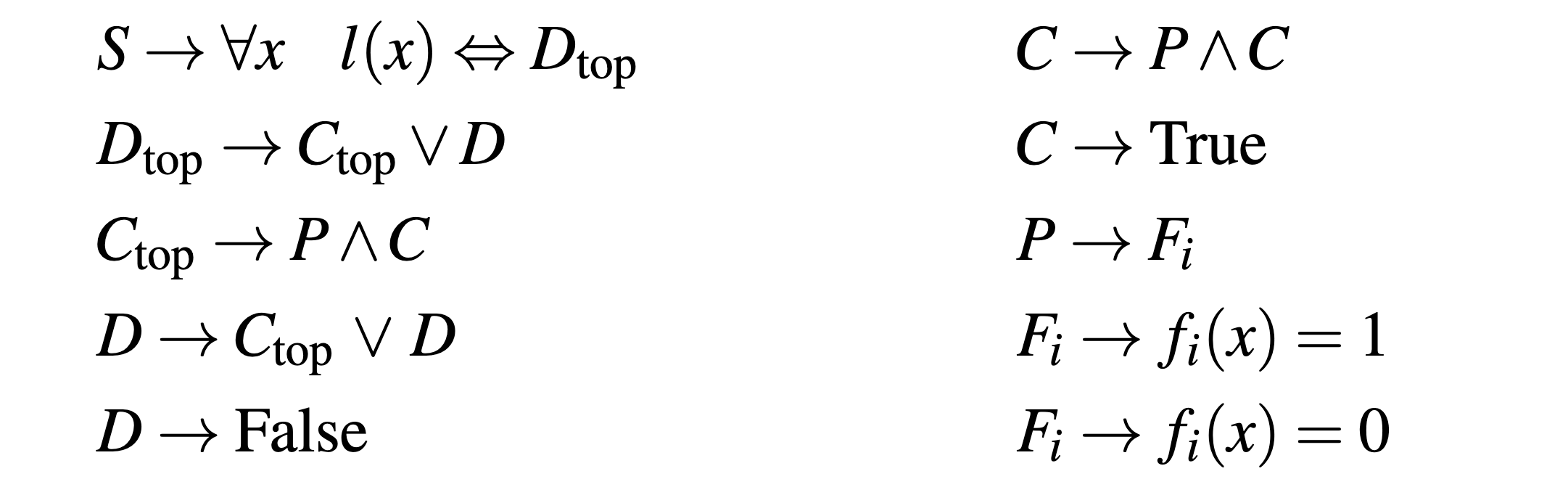}
  \caption{DNF grammar from \protect\citeA{goodman}. \newline $i \in \{1,2,... \text{number of features}\}$  }
  \label{fig:grammar}
\end{figure}

To define our target inductive bias, we use the prior from the Bayesian model created by \citeA{goodman}. This prior is defined using the context-free grammar (CFG) shown in Figure \ref{fig:grammar}.\footnote{\citeA{goodman} denote their CFG in a different way that does not include the rules subscripted with \textit{top}. However, we reran their experiments using the package Fleet \cite{yang2022one} and were able to replicate their results only when including the \textit{top} rules. We therefore infer that \citeauthor{goodman}'s grammar included something akin to the \textit{top} rules that was omitted from their diagrams for conciseness.} CFGs are best known for being used to characterize the syntax of natural languages and to generate sentences in a language; in the case of \citeauthor{goodman}'s model, the CFG instead generates concept definitions. 

The derivation of a concept definition starts with the symbol $S$. This symbol is then recursively expanded using the rules of the grammar until it has yielded a complete expression. For instance, the $S$ would first be expanded into $\forall x ~ l(x) \Leftrightarrow D_{\rm top}$. We would then expand $D_{\rm top}$ to yield $\forall x ~ l(x) \Leftrightarrow C_{\rm top} \lor D$, after which $C_{\rm top}$ and $D$ would each need to be expanded. 
Eventually this would produce a definition such as $\forall x ~ l(x) \Leftrightarrow f_3(x) = 1 \land f_2(x) = 0 \land \text{True} \lor \text{False}$. This expression defines the concept characterized by feature 3 having a value of 1 and feature 2 having a value of 0 (other parts of the definition---$\forall x ~ l(x)$, True, False---are included for formal reasons but do not affect the definition).

In order to make these derivations probabilistic, each rule in the grammar has a probability attached to it, and expansions are then sampled according to those probabilities. %---i.e., the CFG is made into a PCFG.  

\subsection{Neural network architectures for few-shot learning} Although standard neural network systems are not effective at learning from few examples, some alternative model architectures have been proposed that are tailored for this purpose. Examples include Matching Networks \cite{vinyals2016matching} and Prototypical Networks \cite{snell2017prototypical}. Our approach differs from these in modifying the training regime rather than  the neural network's architecture or processing mechanisms.

\subsection{Meta-learning} 

Besides specially-designed architectures, the other main approach for enabling neural networks to learn from few examples is meta-learning. 
In this work, we adopt a type of meta-learning called 
Model-Agnostic Meta-Learning (MAML) \cite{maml}. MAML is a machine learning technique that focuses on training a model in such a way that it can quickly adapt to new, unseen tasks with minimal data. MAML aims to learn an effective initialization for a neural network so that fine-tuning on a new task requires fewer iterations or samples. The model is trained on a diverse set of tasks (from the same task distribution) during a meta-training phase, learning the set of initial parameters. Upon encountering an unseen task, the model's learned initialization can be rapidly fine-tuned with a small amount of data from a new task. Meta-learning extracts common structure from a set of tasks, resulting in a system that has inductive biases aligned with those tasks. 

MAML finds parameters $\theta$ that minimize the summed loss over tasks:
\begin{equation}
\underset{\theta}{\text{argmin}}\sum_{i=1}^{N} \mathcal{L}_{\text{val}} \left( \theta - \alpha \nabla_{\theta} \mathcal{L}_{\text{train}}(\theta) \right),
\end{equation}
where
\( \theta \) represents the model parameters, \( N \) is the number of tasks in the meta-training set, \( \mathcal{L}_{\text{val}} \) is the validation loss used to update the parameters, and \( \alpha \) is the step size or learning rate for the inner loop optimization.
 
Prior work has established that meta-learning is a promising tool for building models of human cognition. In some cases, it is used to model ways in which humans themselves meta-learn \cite{griffiths2019doing,wang2021meta,kumar2021metalearning}. In other cases---as in our work---meta-learning is used as a tool to create neural networks that have human-like inductive biases, without making the claim that meta-learning is \textit{how} humans arrived at those biases; this framing has been used to improve how well neural networks can learn symbolic rules \cite{lake2019compositional,mccoy2020universal,lake2023human}. Meta-learning is also powerful for connecting neural networks to other research traditions in cognitive science: \citeA{Binz_Dasgupta_Jagadish_Botvinick_Wang_Schulz_2023} discuss how meta-learning can connect neural networks to rational analysis, and our goal here is to connect neural networks and Bayesian models---an application of meta-learning previously used by \citeA{mccoy2023modeling} in the domain of language, building upon a theoretical connection between MAML and hierarchical Bayesian models \cite{grant2018recasting}. 
%The current paper applies \citeauthor{mccoy2023modeling}'s approach to a new domain, namely concept learning. 
Note that Prior-data Fitted Networks \cite{muller2022transformers} also inject Bayesian priors into neural networks, but they do so in a different manner that uses standard learning rather than meta-learning.

\section{Distilling Priors into a Neural Network}
To distill a Bayesian prior into a neural network, we use the three-step approach illustrated in Figure \ref{fig:method}. We first use a probabilistic model to define the inductive bias that we wish to give to a neural network. 
Specifically, this probabilistic model is the component of a Bayesian model that defines its prior, and it gives a distribution over possible concepts. Internalizing this distribution would help a neural network learn more efficiently by shaping the hypothesis space that it searches over. In order to make this distribution accessible to a neural network, we sample a large number of concepts from the probabilistic model  for use as training data.  We then have a neural network meta-learn from these sampled tasks to give it an inductive bias that matches the prior distribution we started with.
This approach was first used in previous work that modeled language learning \cite{mccoy2023modeling}; the current paper applies the approach in a new domain, namely concept learning.

% \subsection{Probabilistic model}

\subsection{Sampling data}
We sample many concepts from the grammar in Figure \ref{fig:grammar} in order to serve as meta-learning data for a neural network. 
For each concept, we first sample the production probabilities from a Dirichlet distribution (note that, following \citeA{goodman}, the CFG production rules are fixed, but their probabilities are not).
We then sample a concept definition from the grammar with those probabilities.
We must next generate both a training set and test set for this concept. 
The training set contains up to 20 examples, randomly sampled from the 16 possible objects. We use multi-step loss \cite{antoniou2019train}, which has the same effect as providing the learner with training sets of various sizes, and we also allow examples to repeat within one episode. The test set is made of all 16 possible objects. Each example is labeled True if it obeys the rule defining the concept or False otherwise, except that with probability
$e^{-b}/(1 + e^{-b})$ we flip the label of the example (this is how \citeauthor{goodman}'s model incorporates the possibility of outliers). We denote this outlier parameter  $b \in \{1,2,..,8\}$.

The above procedure produces the data for a single concept. We repeat this procedure to produce data for 10,000 concepts, which will serve as the data from which the neural network will meta-learn.

%Since the Bayesian models and PCFGs, we can generate concepts of arbitrary complexity. In this work, we look at logical concepts, whose production rules in the PCFG involve Boolean operators $\neg, \lor, \land$. We conduct two sets of experiments and, for each, we generate 10000 meta-training tasks from the Bayesian model's DNF grammar \cite{goodman}.

\subsection{Inductive bias distillation via meta-learning}  

The neural network internalizes the Bayesian prior by meta-learning through experience with many similar tasks, each corresponding to learning one concept. We framed the concept learning problem as a binary classification task and assumed the desired concept was learned if specific examples were classified correctly. The architecture we used in our experiments is a multi-layer perceptron (MLP). We generated 10,000 episodes for meta-training, 100 for meta-validation, and 100 for meta-testing. 
We start with a baseline MLP which uses hidden size 128 and 5 layers, dropout 0.1, 1 epoch, outer learning rate 0.0005, and inner learning rate 0.1. We also use a modified version which has hidden size 256 and skip connections. We use the term \textbf{prior-trained} network to refer to a network that has undergone this distillation process; a \textbf{standard} network is one that has not undergone this process.

\section{Evaluating the Model}
% TODO: We should definitely discuss ways in which performance is influenced by details of the NN setup that we use. The general point would be that the inductive bias we end up with is not fully determined by the Bayesian prior but is also affected by the nature of the system we distill it into.
% -describe what rational rules paper does + what kind of experiments are there\\

To evaluate the model we examined its performance in accounting for a set of behavioral results from experiments that had previously been used to assess the Rational Rules model \cite{medin&schaffer, medin&schwanenflugel, Shepard1961LearningAM, McKinley}.
These experiments involve learning logical concepts corresponding to objects that have three or four Boolean features. For example, a concept could be represented by ``$f_1(x) = 0$ and $f_3(x)=1$" meaning that the first feature has a value of 0 and the third feature, 1. The human experiments required the participants to classify the objects into binary categories, according to the rule they inferred. We test our model against the Rational Rules model and human data. 

% \subsection{Grammar} We sampled data for meta-training based on the DNF grammar of \citeA{goodman}, which we adapted. Using this grammar, about $90\%$ of sampled rules are just “True” or “False.” We augment the DNF grammar to produce more complex rules, which leads to about $30\%$ of rules being equivalent to “True” or “False.” by adding the following production rules : $D\_top, C\_top$.

% -describe how we sample tasks\\
% We produce the rules for the training tasks in the following way. We first sample the production probabilities for the above grammar from a Dirichlet distribution (note that the CFG productions are fixed, but their probabilities are not). Then we sample a rule from the PCFG with those probabilities. For each rule, we generate training examples and obtain a label True if it obeys the rule and False otherwise. For the test set, we use all $2^{num\_{features}}$ examples. With probability
% $e^{-b}/(1 + e^{-b})$, we flip the label of the example in order to obtain outliers.

\subsection{Category structure of \citeA{medin&schaffer}}
\label{sec:medin&schaffer}

% - exp 1 (table3 from RR)\\
In Table \ref{tab: medinschaffer}, we consider the category structure of \citeA{medin&schaffer}, using  human data from \citeA{Nosofsky1994RuleplusexceptionMO}, and compare the behavior of humans, the Rational Rules model with $b=1$, the prior-trained neural network, and the standard neural network.

\begin{table}
\caption{The category structure of \protect\citeA{medin&schaffer}, with the human data of \protect\citeA{Nosofsky1994RuleplusexceptionMO}, the predictions of the Rational Rules model with $b = 1$, the predictions of the prior-trained MLP with $b = 2$, and the predictions of a standard (non-prior-trained) MLP.}
\label{tab: medinschaffer}
\small
\setlength{\tabcolsep}{5pt} % Adjust the inter-column space
\centering
\begin{tabular}{c c c c c c}
\hline
\textbf{Object}&  \begin{tabular}{@{}c@{}}\textbf{Feature} \\ \textbf{values} \end{tabular}& \textbf{Human}& $\bf{RR_{DNF}}$ &  \begin{tabular}{@{}c@{}}\textbf{Prior} \\ \textbf{trained} \end{tabular} & \textbf{Standard} \\
\hline
A1 & 0001 & 0.77 & 0.82 & 0.71 & 0.52\\
A2 & 0101 & 0.78 & 0.81 & 0.76 & 0.52\\
A3 & 0100 & 0.83 & 0.92 & 0.84 & 0.52\\
A4 & 0010 & 0.64 & 0.61 & 0.69 & 0.52\\
A5 & 1000 & 0.61 & 0.61 & 0.70 & 0.52\\
B1 & 0011 & 0.39 & 0.47 & 0.40 & 0.52\\
B2 & 1001 & 0.41 & 0.47 & 0.45 & 0.52\\
B3 & 1110 & 0.21 & 0.21 & 0.22 & 0.52\\
B4 & 1111 & 0.15 & 0.07 & 0.14 & 0.52\\
T1 & 0110 & 0.56 & 0.57 & 0.56 & 0.52\\
T2 & 0111 & 0.41 & 0.44 & 0.34 & 0.52\\
T3 & 0000 & 0.82 & 0.95 & 0.84 & 0.52\\
T4 & 1101 & 0.40 & 0.44 & 0.41 & 0.52\\
T5 & 1010 & 0.32 & 0.28 & 0.39 & 0.52\\
T6 & 1100 & 0.53 & 0.57 & 0.60 & 0.52\\
T7 & 1011 & 0.20 & 0.13 & 0.19 & 0.52\\
\hline
\end{tabular}
% wrong last column!! those are results for b=2 and they have RR for b=1

\end{table}

Each object has four binary features. An object can therefore be represented by a sequence of four zeroes or ones; e.g., ``0111" means that the object has a value of 0 for feature 1, a value of 1 for feature 2, a value of 1 for feature 3, and a value of 1 for feature 4. Learners (whether humans or computational models) were trained on 9 of the 16 possible objects: they were shown an object's feature values along with a label indicating whether the object belonged to category A or category B. The training examples are the ones labeled ``A" or ``B" in the Object column of Table \ref{tab: medinschaffer}. Participants were then shown all 16 concepts (the 9 they had been trained on, plus the 7 they had not been---labeled T1 through T7). 
% and it was ascertained what probability they assigned to each one belong to category A. 

The last 4 columns of the table show the probability that the human or model assigned category A to a given object. For instance, humans had 0.77 probability of labeling A1 as belonging to category A. There are multiple possible rules that a learner could consider when trained on this dataset. For instance, the learner could learn a complicated rule that captures all of the training data perfectly, or it could learn a rule that misclassifies some of the training data but is simpler, perhaps under the assumption that the misclassified examples are outliers. One example of such a simpler rule would be to assume that objects with a value of 0 for feature 1 are in category A, while those with a value of 1 for feature 1 are in category B. This rule correctly classifies 7 of the 9 training examples. This interplay between multiple possible rules can be seen in the varying probability levels that are assigned to different training examples; e.g., even though learners have seen A1 through A5 as training examples labeled A, not all of them have the same probability of being labeled A by the learners; presumably, this is because some of them are consistent with more of the candidate rules than others.

\begin{figure}
    \centering
    \includegraphics[width=\linewidth]{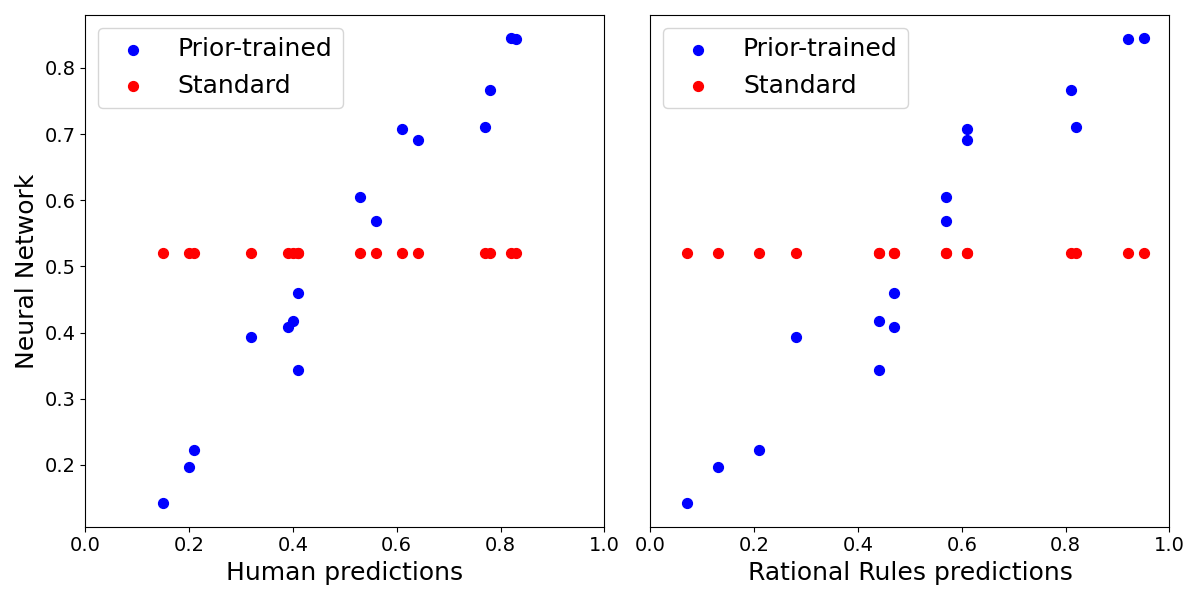}
    \caption{Predictions of prior-trained and standard neural networks vs humans and vs Rational Rules (data from Table~\ref{tab: medinschaffer}).}
    \label{fig:mamlvsstandard}
\end{figure}

% \begin{figure}[ht]
%     \centering
%     \includegraphics[width=\linewidth]{figures/table3_humans.png}  % Replace with the filename of your first figure
%   \end{minipage}%
%   \begin{minipage}{.25\textwidth}
%     \centering
%     \includegraphics[width=\linewidth]{figures/table3_RR.png}  % Replace with the filename of your second figure
%   \end{minipage}
%   \caption{Predictions of prior-trained and randomly initialized neural network vs humans and vs Rational Rules}
%   \label{fig:mamlvsstandard}
% \end{figure}

The baseline architecture with a random initialization obtains $R^2=0$. Using the same model with the meta-learned initialization, we find $R^2 = 0.95$ for the prior-trained neural network and humans and $R^2=0.92$ for the prior-trained neural network and the Rational Rules model, demonstrating that meta-trained models can closely match the human results; $R^2 = 0.98$ between the Rational Rules model and humans.  The specific testing examples as well as the error probability for each model are shown in Table \ref{tab: medinschaffer}. Figure \ref{fig:mamlvsstandard} shows that the prior-trained neural network's predictions are highly correlated with the human predictions, while those of the standard neural network are not.

\subsection{Linear Separability}
% -exp 2(table 4 + fig 5 from RR) - todo remake figure with error bars

\begin{figure*}[ht]
    \centering
    
    \begin{minipage}{0.24\textwidth}
        \centering
        \includegraphics[width=\linewidth]{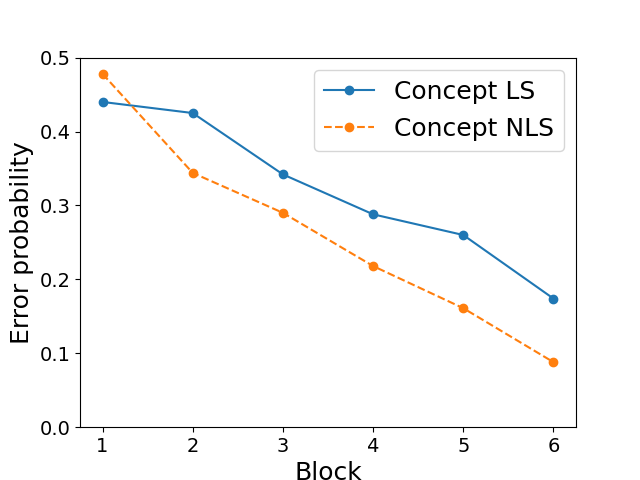} % Replace with the filename of your first plot
        \subcaption*{(a) Humans}
    \end{minipage}%
    \hfill
    \begin{minipage}{0.24\textwidth}
        \centering
        \includegraphics[width=\linewidth]{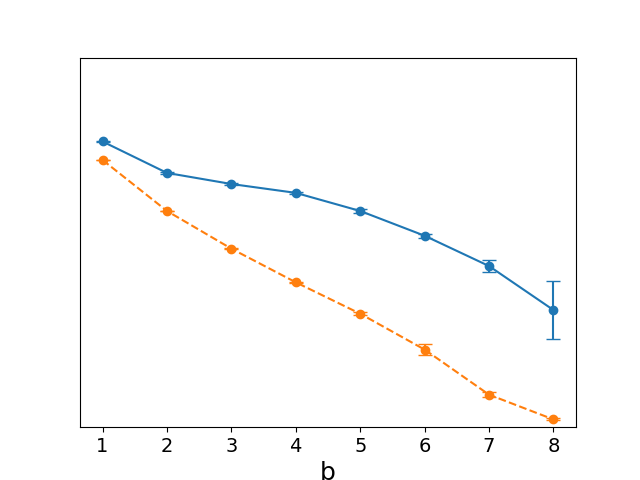} % Replace with the filename of your first plot
        \subcaption*{(b) RR$_\text{DNF}$}
    \end{minipage}%
    \hfill
    \begin{minipage}{0.24\textwidth}
        \centering
        \includegraphics[width=\linewidth]{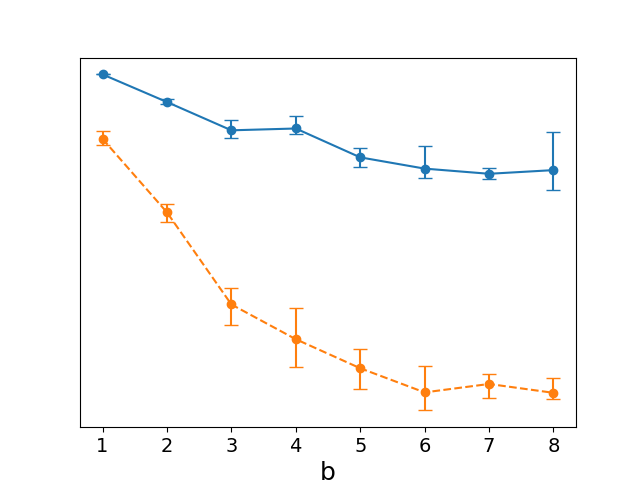} % Replace with the filename of your second plot
        \subcaption*{(c) Prior-trained (varying $b$)}
    \end{minipage}%
    \hfill
    \begin{minipage}{0.24\textwidth}
        \centering
        \includegraphics[width=\linewidth]{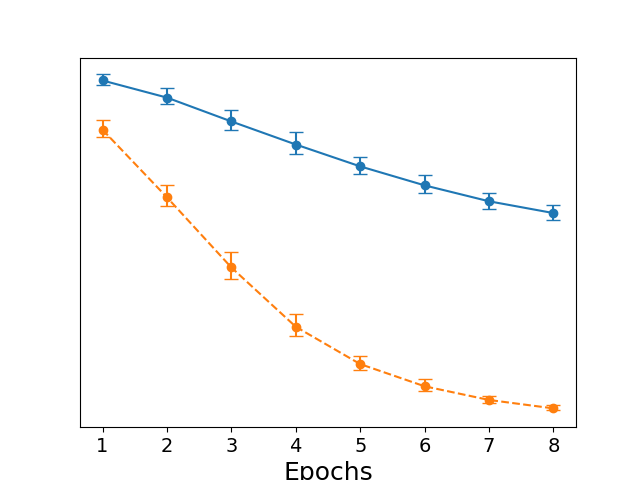} % Replace with the filename of your third plot
        \subcaption*{(d) Prior-trained (varying epochs)}
    \end{minipage}
    
    \caption{Linearly and non-linearly separable concepts. Linearly separable Concept LS was more difficult to learn than Concept NLS, which is not linearly separable. Error probability in: (a) Humans, (b) Rational Rules, (c) Prior-trained modified model varying outlier parameter $b$, (d) Prior-trained baseline model varying number of epochs per episode during inference}
    \label{fig:lsnls}
\end{figure*}

We next consider the two concepts from \citeA{medin&schwanenflugel} which have four binary features and are shown in Table \ref{tab:lsnls}. Concept LS is linearly separable, meaning that it admits a linear discriminant boundary, while Concept NLS is not. Our model predicts that that Concept NLS is easier to learn, in agreement with the human and Rational Rules predictions. In Figure \ref{fig:lsnls} we show the error probability: $1 - {P}(\text{true label} = \text{predicted label})$ in the Rational Rules model and two variations of our model; the output of our model is ${P}(\text{Category A})$. The two setups we study are: the modified model varying the outlier parameter $b \in \{1,2...8\}$ with one epoch per episode at test time and the baseline model meta-trained with $b=1$ varying $N \in \{1,2...8\}$ epochs per episode at test time. Both neural network variants show trends similar to those found with the Rational Rules model, which in turn behaved similarly to humans \cite{goodman}.

\begin{table}[ht]
     \caption{Two concepts in Medin \& Schwanenflugel (1981). Concept LS is linearly separable, Concept NLS is not.}
    \label{tab:lsnls}
    \centering
    \begin{minipage}{0.45\linewidth} % Adjust the width as needed
        \centering
        \begin{tabular}{cc}
            \multicolumn{2}{c}{\textbf{Concept LS}} \\
            \hline
            Category A & Category B \\
            \hline
            1000 & 0111 \\
            0001 & 1000 \\
            0110 & 1001 \\
            \hline
        \end{tabular}
    \end{minipage}%
    \hfill\begin{minipage}{0.45\linewidth} % Adjust the width as needed
        \centering
        \begin{tabular}{cc}
            \multicolumn{2}{c}{\textbf{Concept NLS}} \\
            \hline
            Category A & Category B \\
            \hline
            0011 & 1111 \\
            1100 & 1010 \\
            0000 & 0101 \\
            \hline
        \end{tabular}
    \end{minipage}
\end{table}

\begin{table}[bh!]
    \caption{The six three-feature concepts with four positive and four negative examples, studied by \protect\citeA{Shepard1961LearningAM}.}
    \label{tab:sixconcepts}
    \centering
    \begin{tabular}{cccccc}
        \hline
        \textbf{I} & \textbf{II} & \textbf{III} & \textbf{IV} & \textbf{V} & \textbf{VI} \\
        \hline
        +000 & +000 & +000 & +000 & +000 & +000 \\
        +001 & +001 & +001 & +001 & +001 & -001 \\
        +010 & -010 & +010 & +010 & +010 & -010 \\
        +011 & -011 & -011 & -011 & -011 & +011 \\
        -100 & -100 & -100 & +100 & -100 & -100 \\
        -101 & -101 & +101 & -101 & -101 & +101 \\
        -110 & +110 & -110 & -110 & -110 & +110 \\
        -111 & +111 & -111 & -111 & +111 & -111 \\
        \hline
    \end{tabular}

\end{table}

\begin{figure}[ht]
  \centering
  \begin{minipage}{.25\textwidth}
    \centering
    \includegraphics[width=\linewidth]{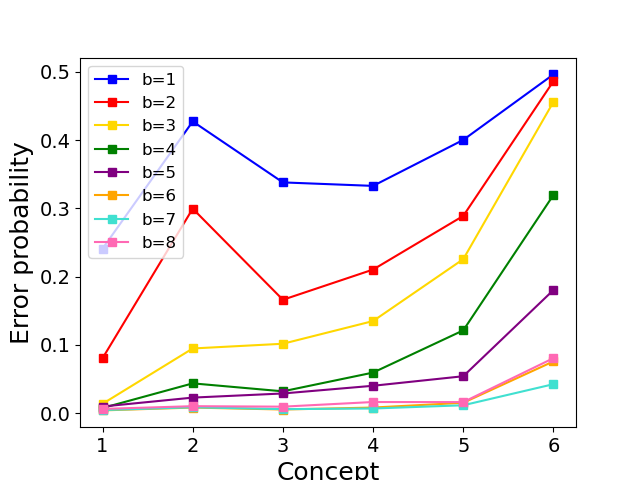}  % Replace with the filename of your first figure
    \subcaption*{(a)}
  \end{minipage}%
  \begin{minipage}{.25\textwidth}
    \centering
    \includegraphics[width=\linewidth]{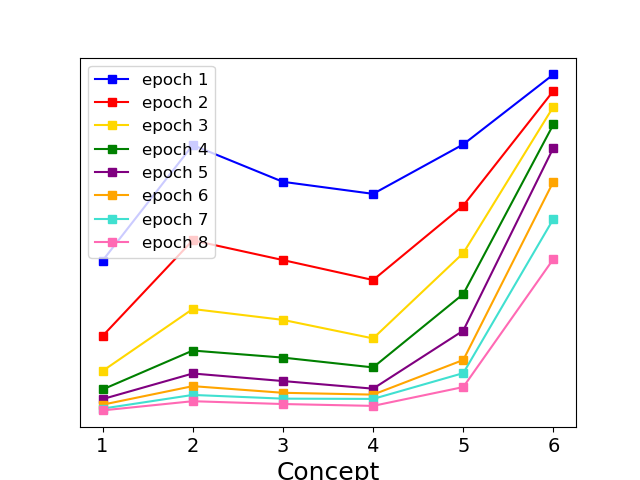}  % Replace with the filename of your second figure
    \subcaption*{(b)}
  \end{minipage}
  \caption{Error probability of the prior-trained model in learning concepts I-VI in \protect\citeA{Shepard1961LearningAM}. (a) Prior-trained modified model varying outlier parameter $b$. (b) Prior-trained baseline model varying number of epochs per episode.}
  \label{fig:sixconcepts}
\end{figure}

\subsection{Shepard, Hovland and Jenkins (1961)}
% -exp 3 (table 5 from RR)

\citeA{Shepard1961LearningAM} compared difficulty in learning the six concepts shown in Table \ref{tab:sixconcepts}. The concepts have  three Boolean features and are divided into four positive and four negative examples. 
The human performance in learning these concepts indicates the following ranking of difficulty in learning six concepts shown in Table \ref{tab:sixconcepts}: $I < II < III = IV = V < VI$.  The Rational Rules error rates are $0\%,17\%, 24\%,24\%,25\%,48\%$, for $b=3$. The ranking is consistent for all values of $b$ other than $b=1$, where there is an inversion: $II > III$. Our prior-trained model exhibits the same trend, including the inversion $II > III$ for $b=1$ and $b=2$, which we show in Figure \ref{fig:sixconcepts} for the modified model meta-trained varying $b$ and for the baseline model meta-trained with $b=1$ and varying the number of epochs per episode at test time.

\subsection{\citeA{medin82}}
This experiment follows the same setup as our experiment based on  \citeA{medin&schaffer}, meta-training our modified model with $b=1$ and $b=7$. In Table \ref{tab:medinetal}, we test our model on the category structure of \citeA{medin82}. The $RR_{DNF}$ model explains most of the variance in human judgments in the final stage of learning: $R^2 = 0.95$ when $b = 7$ and correlation with human judgments after one training block is $R^2 = 0.69$ when $b = 1$. Our model has $R^2 = 0.56$ when $b=7$ and $R^2 = 0.66$ when $b=1$; see the Discussion for a potential explanation for why the $b = 7$ result does a poorer job of replicating $RR_{DNF}$ than our other results do.

\begin{table}[t]
    \caption{The category structure of \protect\citeA{medin82}, with initial and final block mean human responses of \protect\citeA{McKinley}, and the predictions of the Rational Rules model and our model at $b = 1$ and $b = 7$.}
    \label{tab:medinetal}
\small % Adjust the font size
\setlength{\tabcolsep}{1pt} % Adjust the inter-column space
\renewcommand{\arraystretch}{1.2} % Adjust the row height if needed
    \centering
    \begin{tabular}{cc|ccc|ccc}
    \hline
         \textbf{Object}&  \begin{tabular}{@{}c@{}}\textbf{Feature} \\ \textbf{values} \end{tabular}&  \begin{tabular}{@{}c@{}}\textbf{Human} \\  \begin{tabular}{@{}c@{}}\textbf{initial} \\ \textbf{block}\end{tabular}\end{tabular}&  \begin{tabular}{@{}c@{}}$\bf{RR_{DNF}}$ \\ $\bf{b=1}$\end{tabular}&  \begin{tabular}{@{}c@{}}\textbf{MAML} \\ $\bf{b=1}$\end{tabular}&  \begin{tabular}{@{}c@{}}\textbf{Human} \\  \begin{tabular}{@{}c@{}}\textbf{final} \\ \textbf{block}\end{tabular}\end{tabular}&  \begin{tabular}{@{}c@{}}$\bf{RR_{DNF}}$ \\  $\bf{b=7}$\end{tabular}& \begin{tabular}{@{}c@{}}\textbf{MAML} \\ $\bf{b=7}$\end{tabular}\\
         \hline
         A1&  1111&  0.64&  0.84&  0.84&  0.96&  1& 0.98\\
         A2&  0111&  0.64&  0.54&  0.67&  0.93&  1& 0.97\\
         A3&  1100&  0.66&  0.84&  0.83&  1&  1& 0.98\\
         A4&  1000&  0.55&  0.54&  0.66&  0.96&  0.99& 0.96\\
         B1&  1010&  0.57&  0.46&  0.32&  0.02&  0& 0.03\\
         B2&  0010&  0.43&  0.16&  0.15&  0&  0& 0.02\\
         B3&  0101&  0.46&  0.46&  0.31&  0.05&  0.01& 0.03\\
         B4&  0001&  0.34&  0.16&  0.15&  0&  0& 0.02\\
         T1&  0000&  0.46&  0.2&  0.22&  0.66&  0.56& 0.14\\
         T2&  0011&  0.41&  0.2&  0.26&  0.64&  0.55& 0.32\\
         T3&  0100&  0.52&  0.5&  0.45&  0.64&  0.57& 0.3\\
         T4&  1011&  0.5&  0.5&  0.48&  0.66&  0.56& 0.38\\
         T5&  1110&  0.73&  0.8&  0.72&  0.36&  0.45& 0.66\\
         T6&  1101&  0.59&   0.8&  0.74&  0.36&  0.44& 0.79\\
         T7&  0110&  0.39&  0.5&  0.49&  0.27&  0.44& 0.53\\
         T8&  1001&  0.46&  0.5&  0.51&  0.3&  0.43& 0.63\\
         \hline
    \end{tabular}

\end{table}

\section{Discussion}

Humans can learn logical concepts rapidly, but it is natural to expect that this task will be difficult for artificial neural networks. First, these concepts are defined using discrete, symbolic, compositional rules, but neural networks are not a natural fit for symbolic domains \cite{fodor1988connectionism}. Second, in the human experiments that we considered, such concepts were learned from small numbers of examples, yet standard neural networks usually require large quantities of training data.
Indeed, we found that standard neural networks did a poor job at acquiring the concepts we considered. Nonetheless, we also found that inductive bias distillation made it possible for neural networks to perform well at learning logical concepts: after we distilled a structured prior probability distribution into a neural network, it was able to learn logical concepts from few examples and in ways that aligned closely with human results across several experiments.
These results therefore show that it is possible to develop connectionist instantiations of probabilistic models.

%-results interpretation \\

\subsection{Potential additional biases}
The prior that we distilled into a neural network encodes a bias for simplicity: rules that can be expressed with a short logical description have a higher prior probability than rules that require a long description.
Such a bias is well-supported by human experiments \cite{feldman2000minimization,neisser1962hierarchies,goodman}, but it is far from the only bias involved in human concept learning. 
For instance, people also display a shape bias \cite{landau1988importance}, a whole-object bias \cite{markman1994constraints}, a basic-level bias \cite{rosch1976basic}, and a mutual exclusivity bias \cite{markman1988children}. 
One direction for future work would be to incorporate these additional biases into our model, which could be done by augmenting the probabilistic model from which we sampled concepts such that the distribution of concepts that it produces reflects these biases. For instance, a shape bias could be instantiated by introducing an asymmetry between features so that concepts based on the shape feature would be sampled more often than concepts based on other features.

\subsection{Modeling the timecourse of learning}
When our neural network model learns a concept, it does so incrementally, updating its parameters after each example it encounters. 
This incrementality means that the neural network can model the timecourse of learning more naturally than the Rational Rules model, which considers the entire set of training examples at once, such that modeling multiple points during the course of learning requires creating multiple separate instances of the model.

%The learning process is captured better by a neural network than a Bayesian model, as the neural network is closer to an algorithmic model. In our experiments, this is realized by starting with a prior-trained model and then vary the number of epochs when test examples are shown, instead of training many different models by varying the outlier parameter $b$.

As one example,
in some of the human experiments \cite{medin&schwanenflugel, Shepard1961LearningAM}, participants were trained on the category using a blocked learning paradigm: each example in the training set was presented once per block, and blocks were repeated multiple times. 
To replicate this experiment with the Rational Rules model, a different instance of the model must be created for each step in the learning process. Specifically, as shown in Figure \ref{fig:lsnls}b, successive steps are modeled by increasing the $b$ hyperparameter, though the same result could also be achieved by fitting each instance to differing numbers of copies of training examples: the Rational Rules model with outlier parameter $b$ presented with $N$ identical blocks of examples is equivalent to the model presented with only one block, but with parameter $b' = b \cdot N$. 

Like Rational Rules, the neural network can accommodate differences in $b$ (Figure \ref{fig:lsnls}c), but unlike Rational Rules it also accommodates an approach in which the same copy of the model is incrementally trained on multiple repeats of the training set (Figure \ref{fig:lsnls}d)---an approach that better captures incremental concept learning in humans.

\subsection{Challenges of learning a prior from data}
One shortcoming of our approach arises from the fact that the neural network model learns its inductive bias from data. 
Therefore, the neural network is likely to have approximated this bias very well for the scenarios that were well-represented in the data, but the approximation may be poor in rarer scenarios. 
In our case, the frequency with which the network encountered a given concept was equal to that concept's prior probability, so the degree to which our model approximates the Rational Rules model is likely to decrease when a concept with a low prior probability is involved, since our model has seen few examples of how Rational Rules would generalize for such concepts.

We had one result where the neural network indeed provided a poor fit to Rational Rules, namely the $b=7$ setting in Table \ref{tab:medinetal}. The low performance in this case is likely due to the fact that the rule to be learned, namely $(f_3 = 1 \land f_4 = 1) \lor (f_3 = 0 \land f_4 = 0)$, has a low prior probability (because its description needs to include a large number of feature specifications) and thus is outside the space in which the neural network's parameters are well-estimated. In contrast, the $b=1$ setting in that table involves the same data but with a higher outlier probability; increasing the outlier probability leads Rational Rules to ignore some of the training examples as outliers in order to learn a simpler (i.e., higher-prior) rule, and having a high prior for the rule to be learned facilitates a stronger performance from the neural network model. 

Enabling neural networks to generalize to low-probability scenarios is a challenging problem. Meta-learning is an active area of research in machine learning, and technological advances in this area may provide ways to overcome this problem, but for now it remains an important limitation for data-driven approaches to imparting inductive biases.

\subsection{Conclusion}

While Bayesian models excel in certain aspects of concept learning, their traditional reliance on explicitly symbolic representations limits their implementational and algorithmic plausibility. On the other hand, neural networks do not require such representations but struggle to learn from small numbers of examples. Our approach bridges this gap by distilling probabilistic models into neural networks, leveraging the strengths of both paradigms. This integration enables neural networks to learn structured, compositional concepts from limited numbers of examples, providing new avenues for developing theories of concept learning that merge insights from Bayesian models and neural networks.

\subsection*{Acknowledgments}

This work was supported by grant N00014-23-1-2510 from the Office of Naval Research.
RTM was supported in part by the NSF SBE Postdoctoral Research Fellowship under Grant No.\ 2204152.

\nocite{ChalnickBillman1988a}
\nocite{Feigenbaum1963a}
\nocite{Hill1983a}
\nocite{OhlssonLangley1985a}
% \nocite{Lewis1978a}
\nocite{Matlock2001}
\nocite{NewellSimon1972a}
\nocite{ShragerLangley1990a}

\bibliographystyle{apacite}

\setlength{\bibleftmargin}{.125in}
\setlength{\bibindent}{-\bibleftmargin}

\bibliography{CogSci_Template}

\end{document}